\documentclass[conference,a4paper]{IEEEtran}
\IEEEoverridecommandlockouts

\usepackage[hidelinks]{hyperref}
\usepackage[cmex10]{amsmath}
\usepackage{amssymb,amsfonts}
\usepackage{dblfloatfix}

\usepackage[ruled,vlined]{algorithm2e}
\usepackage{graphicx}
\graphicspath{{Figures/PDF/}{Figures/PNG/}}
\usepackage{graphicx}    
\usepackage{subcaption}  
\usepackage{booktabs}
\usepackage[table]{xcolor}
\usepackage{booktabs}
\usepackage{siunitx}
\usepackage[numbers,compress]{natbib}
\usepackage{texnames}
\usepackage{bm,bbm}
\usepackage{orcidlink}

\begin{document}

\title{A Benchmark Study of Segmentation Models and Adaptation Strategies for Landslide Detection from Satellite Imagery}

\author{	\IEEEauthorblockN{Md Kowsher}
	\IEEEauthorblockA{\textit{University of Central Florida}\\
		md.kowsher@ucf.edu}
	\and
	\IEEEauthorblockN{Weiwei Zhan}
	\IEEEauthorblockA{\textit{University of Central Florida}\\
		weiwei.zhan@ucf.edu}
	\and
	\IEEEauthorblockN{Chen Chen}
	\IEEEauthorblockA{\textit{University of Central Florida}\\
		chen.chen@ucf.edu}
}

\maketitle
\begin{abstract}
	Landslide detection from high-resolution satellite imagery is a critical task for disaster response and risk assessment, yet the relative effectiveness of modern segmentation architectures and fine-tuning strategies for this problem remains insufficiently understood. In this work, we present a systematic benchmarking study of convolutional neural networks, transformer-based segmentation models, and large pre-trained foundation models for landslide detection. Using the Globally Distributed Coseismic Landslide Dataset (GDCLD) dataset, we evaluate representative CNN- and transformer-based segmentation models alongside large pre-trained foundation models under consistent training and evaluation protocols. In addition, we compare full fine-tuning with parameter-efficient fine-tuning methods, including LoRA and AdaLoRA, to assess their performance–efficiency trade-offs. Experimental results show that transformer-based models achieve strong segmentation performance, while parameter-efficient fine-tuning reduces trainable parameters by up to 95\% with comparable accuracy to full fine-tuning. We further analyze generalization under distribution shift by comparing validation and held-out test performance.

\end{abstract}


\section{Introduction}

Landslides are among the most destructive natural hazards, causing substantial loss of life, infrastructure damage, and long-term socio-economic impacts worldwide \cite{godt2022national, kirschbaum2015spatial, mirus2020landslides, handwerger2019widespread, sato2025variation}. Their occurrence is closely linked to complex interactions among terrain morphology, geological conditions, vegetation cover, precipitation, and seismic activity. Timely and accurate landslide detection is therefore critical for disaster response, risk assessment, and mitigation planning. With the increasing availability of high-resolution satellite imagery, semantic segmentation has become a central paradigm for large-scale, automated landslide mapping in remote sensing.

Recent years have witnessed rapid progress in deep learning–based landslide detection, evolving from convolutional neural networks (CNNs) to transformer-based architectures and large pre-trained foundation models \cite{chen2025harnessing,li2025landslidehazardmappinggeospatial}. CNN-based models such as U-Net \cite{ronneberger2015u} variants and DeepLab \cite{chen2019rethinking} have demonstrated strong performance by exploiting local spatial patterns, while transformer-based models further improve global context modeling and multi-scale feature representation. More recently, general-purpose foundation models—pre-trained on massive, diverse datasets—have shown promise for transfer learning across domains \cite{szwarcman2025prithvi}. However, their effectiveness for geospatial hazards such as landslides, particularly under domain shift and limited labeled data, remains insufficiently understood.

Despite the growing number of reported models, several fundamental questions remain open for the remote sensing community. First, it is unclear how general-purpose vision foundation models compare with widely used CNN- and transformer-based segmentation architectures when applied to landslide detection. Second, while full fine-tuning is commonly adopted, it is computationally expensive and memory intensive, raising the question of whether parameter-efficient fine-tuning (PEFT) strategies—such as Low-Rank Adaptation (LoRA) \cite{hu2022lora} and AdaLoRA \cite{zhang2023adalora}—can achieve comparable performance with significantly reduced training cost. Third, the generalization ability of different model families under distribution shift remains underexplored, particularly when comparing validation performance with held-out test data.

Motivated by these gaps, this paper presents a comprehensive benchmarking study of CNN-based models, transformer-based segmentation architectures, and foundation models for landslide detection. Using the \textit{Globally Distributed Coseismic Landslide} Dataset (GDCLD) \cite{fang2024globally}, we systematically evaluate:
(1) representative CNN- and transformer-based segmentation models versus foundation models;
(2) full fine-tuning versus parameter-efficient fine-tuning strategies; and
(3) generalization across validation and test splits exhibiting distribution shift.

Rather than proposing a new model, our goal is to provide clear empirical evidence that informs model selection and adaptation strategies for landslide segmentation. Our experimental results show that transformer-based models, particularly SegFormer \cite{xie2021segformer} variants, achieve strong overall segmentation performance, while foundation models such as SAM tend to favor higher recall at the cost of precision. Importantly, parameter-efficient fine-tuning reduces the number of trainable parameters by approximately 95\% while achieving performance comparable to full fine-tuning. We further quantify validation-to-test performance degradation to highlight the impact of distribution shift and discuss practical implications for scalable landslide detection in real-world remote sensing applications.

\section{Background}
\label{sec:background}

Semantic segmentation for landslide detection has evolved from traditional convolutional neural networks to modern transformer-based architectures. In this section, we provide background on the baseline methods and foundation models employed in our study.

\subsection{CNN-based Segmentation Models}

\textbf{U-Net} \cite{ronneberger2015u} introduced the encoder-decoder architecture with skip connections, enabling precise localization through the concatenation of low-level and high-level features. The symmetric structure has become foundational for medical and remote sensing image segmentation.

\textbf{ResU-Net} extends U-Net by incorporating residual connections from ResNet-50 \cite{diakogiannis2020resunet} as the encoder backbone, addressing the vanishing gradient problem and enabling deeper network training while preserving spatial information through skip connections.

\textbf{DeepLabV3} \cite{chen2019rethinking} employs atrous spatial pyramid pooling (ASPP) to capture multi-scale contextual information. By applying dilated convolutions at multiple rates, DeepLabV3 handles objects at various scales without sacrificing resolution.

\textbf{HRNet} (High-Resolution Network) \cite{wang2020deep} maintains high-resolution representations throughout the network by connecting multi-resolution subnetworks in parallel, making it particularly effective for tasks requiring precise spatial localization.

\subsection{Transformer-based Segmentation Models}

\textbf{UPerNet} \cite{xiao2018unified} combines a Vision Transformer (ViT-B16) backbone with a unified perceptual parsing framework, leveraging pyramid pooling modules to aggregate multi-scale features for dense prediction tasks.

\textbf{SwinU-Net} \cite{cao2022swin} integrates the Swin Transformer's shifted window mechanism with the U-Net architecture, enabling efficient computation of self-attention within local windows while maintaining cross-window connections.

\textbf{SegFormer} \cite{xie2021segformer} represents a state-of-the-art transformer architecture designed specifically for semantic segmentation. It employs a hierarchical Mix Transformer (MiT) encoder that generates multi-scale features without positional encodings, paired with a lightweight All-MLP decoder. The MiT encoder uses overlapping patch embeddings and efficient self-attention to capture both local and global context. SegFormer-B4 and SegFormer-B5 variants contain approximately 64M and 85M parameters, respectively, offering different trade-offs between computational cost and representational capacity.

\subsection{Foundation Models for Segmentation}

\textbf{Segment Anything Model (SAM)} \cite{kirillov2023segment} is a promptable foundation model developed by Meta AI, pre-trained on the SA-1B dataset containing over 1 billion masks from 11 million images. SAM employs a ViT-Base encoder with a prompt-based mask decoder, designed for zero-shot generalization across diverse segmentation tasks. The model accepts various prompt types including points, bounding boxes, and masks to guide segmentation.

\subsection{Parameter-Efficient Fine-Tuning}

\textbf{Low-Rank Adaptation (LoRA)} \cite{hu2022lora} introduces trainable low-rank decomposition matrices into transformer layers while keeping pre-trained weights frozen. For a pre-trained weight matrix $W_0 \in \mathbb{R}^{d \times k}$, LoRA constrains the update by representing it as:
\begin{equation}
W = W_0 + \Delta W = W_0 + BA
\end{equation}
where $B \in \mathbb{R}^{d \times r}$, $A \in \mathbb{R}^{r \times k}$, and the rank $r \ll \min(d, k)$. This significantly reduces the number of trainable parameters while maintaining adaptation capability.

\textbf{Adaptive Low-Rank Adaptation (AdaLoRA)} \cite{zhang2023adalora} extends LoRA by dynamically allocating the parameter budget among weight matrices based on their importance. AdaLoRA parameterizes incremental updates in singular value decomposition form and prunes less important singular values during training, achieving better performance with similar parameter budgets.

\section{Methodology}
\label{sec:methodology}

\subsection{Problem Formulation}

Given a satellite image $I \in \mathbb{R}^{H \times W \times 3}$, our objective is to predict a binary segmentation mask $M \in \{0, 1\}^{H \times W}$ where pixels belonging to landslide regions are labeled as 1 and background pixels as 0. We address this as a binary semantic segmentation task with significant class imbalance, where landslide pixels constitute approximately 23\% of the total pixels in our dataset.

\subsection{Model Architectures}

\subsubsection{SegFormer Fine-tuning}

We fine-tune SegFormer-B4 and SegFormer-B5 models pre-trained on the ADE20K dataset \cite{zhou2017scene}. The models are initialized from \texttt{nvidia/segformer-b4-finetuned-ade-512-512} and \texttt{nvidia/segformer-b5-finetuned-ade-640-640} checkpoints, respectively. The classification head is replaced with a new decoder head configured for binary segmentation (2 classes: background and landslide).

The MiT encoder produces hierarchical feature maps at resolutions $\{\frac{1}{4}, \frac{1}{8}, \frac{1}{16}, \frac{1}{32}\}$ of the input size. These multi-scale features are fused by the MLP decoder through a series of linear projections and upsampling operations:
\begin{equation}
\hat{F}_i = \text{Linear}(C_i, C)(F_i), \quad i \in \{1, 2, 3, 4\}
\end{equation}
where $F_i$ represents features at scale $i$, and $C$ is the unified channel dimension.

\subsubsection{SAM Adaptation}

For SAM, we employ the ViT-Base variant (\texttt{facebook/sam-vit-base}) with the image encoder frozen during training. We utilize bounding box prompts derived from ground truth masks to guide the mask decoder. The prompt encoder converts bounding box coordinates into positional embeddings that condition the mask prediction.

\subsubsection{Parameter-Efficient Fine-tuning with LoRA and AdaLoRA}

To enable efficient adaptation of foundation models, we apply LoRA and AdaLoRA to the transformer layers. For both methods, we have the following configurations:
\begin{itemize}
    \item Rank $r = 32$
    \item Scaling factor $\alpha = 32$ (effective scaling: $\alpha/r = 1.0$)
    \item Target modules: query and value projection matrices in self-attention layers
    \item Dropout: 0.1
\end{itemize}

For AdaLoRA specifically, we set the initial rank budget to 32 per layer with target rank of 16, allowing the algorithm to adaptively prune less important components during training.

\subsection{Loss Functions}

\subsubsection{Weighted Cross-Entropy Loss}

For SegFormer models, we employ weighted cross-entropy loss to address class imbalance:
\begin{equation}
\mathcal{L}_{WCE} = -\frac{1}{N}\sum_{i=1}^{N} w_{y_i} \log(p_{y_i})
\end{equation}
where $w = [1.0, 3.4]$ represents class weights for background and landslide respectively, computed based on inverse class frequency.

\subsubsection{Binary Cross-Entropy with Logits}

For SAM, we use binary cross-entropy with logits loss with positive class weighting:
\begin{equation}
\small
\mathcal{L}_{BCE} = -\frac{1}{N}\sum_{i=1}^{N} [w_p \cdot y_i \log(\sigma(z_i)) + (1-y_i)\log(1-\sigma(z_i))]
\end{equation}
where $w_p = 3.4$ is the positive weight and $\sigma$ denotes the sigmoid function.

\subsection{Data Augmentation}

During training, we apply color jitter augmentation with the following parameters: Brightness: $\pm 0.25$, Contrast: $\pm 0.25$, Saturation: $\pm 0.25$, and Hue: $\pm 0.1$.

No geometric augmentations (rotation, flipping) are applied to preserve the natural orientation of satellite imagery and landslide morphology.



\begin{table}[h]
\centering
\caption{Training hyperparameters for different model configurations.}
\label{tab:hyperparams}
\resizebox{\columnwidth}{!}{%
\begin{tabular}{@{}lcccc@{}}
\toprule
\textbf{Hyperparameter} & \textbf{SegFormer} & \textbf{SAM} & \textbf{LoRA} & \textbf{AdaLoRA} \\
\midrule
Optimizer & AdamW & AdamW & AdamW & AdamW \\
Learning Rate & $1 \times 10^{-4}$ & $1 \times 10^{-4}$ & $2 \times 10^{-4}$ & $2 \times 10^{-4}$ \\
Weight Decay & 0.01 & 0.01 & 0.01 & 0.01 \\
Batch Size & 16 & 8 & 16 & 16 \\
Epochs & 4 & 2 & 5 & 5 \\
LR Scheduler & Linear & Linear & Cosine & Cosine \\
Warmup Ratio & 0.1 & 0.1 & 0.05 & 0.05 \\
Input Size & $512 \times 512$ & $256 \times 256$ & $512 \times 512$ & $512 \times 512$ \\
LoRA Rank & - & - & 32 & 32 \\
LoRA Alpha & - & - & 32 & 32 \\
\bottomrule
\end{tabular}%
}
\end{table}

\subsection{Evaluation Metrics}

We evaluate all models using standard semantic segmentation metrics computed at the macro level across both classes:

\textbf{Precision}: The ratio of correctly predicted positive pixels to all predicted positive pixels.

\textbf{Recall}: The ratio of correctly predicted positive pixels to all actual positive pixels.

\textbf{F1-Score}: The harmonic mean of precision and recall.

\textbf{Mean Intersection over Union (mIoU)}: The average IoU across all classes:
\begin{equation}
\text{mIoU} = \frac{1}{C}\sum_{c=1}^{C} \frac{TP_c}{TP_c + FP_c + FN_c}
\end{equation}
where $C=2$ is the number of classes, and $TP$, $FP$, $FN$ denote true positives, false positives, and false negatives respectively.

\section{Experiments}
\label{sec:experiments}

\subsection{Dataset}

We conduct experiments on the Globally Distributed Coseismic Landslide Dataset (GDCLD) \cite{fang2024globally}, which contains high-resolution satellite imagery with pixel-wise annotations from nine earthquake-triggered landslide events. The dataset is split into \textbf{Training set}: 11,162 images, \textbf{Validation set}: 4,459 images, and \textbf{Test set}: 1,091 images.

Images are provided in GeoTIFF format with corresponding binary masks. The class distribution exhibits significant imbalance with approximately 77\% background pixels and 23\% landslide pixels, motivating our use of weighted loss functions.

\subsection{Implementation Details}

All experiments are conducted on NVIDIA H100 GPUs with 80GB memory. We use PyTorch 2.0 with mixed-precision training (FP16) for computational efficiency. The HuggingFace Trainer API is employed for training management with gradient accumulation when necessary for larger batch sizes.

For LoRA and AdaLoRA experiments, we use the PEFT (Parameter-Efficient Fine-Tuning) library. The adapters are applied to the query and value projection matrices in all transformer layers of the encoder. Table \ref{tab:hyperparams} summarizes the hyperparameters. 

\subsection{Results}

\subsubsection{Validation Dataset Results}

Table~\ref{tab:validation} presents the comparison of all methods on the GDCLD validation dataset \cite{fang2024globally}. Among CNN-based methods, SwinU-Net achieves the highest mIoU of 83.68\%, followed by UPerNet at 81.97\%. The SegFormer baseline achieves 85.06\% mIoU, demonstrating the effectiveness of transformer architectures for this task.

Our fine-tuned models show competitive performance. SegFormer-B5 achieves 82.95\% mIoU with the highest recall (94.38\%) among all models, indicating strong landslide detection capability. SegFormer-B4 follows closely with 81.17\% mIoU. Foundation model SAM, despite being designed for general-purpose segmentation, achieves 73.58\% mIoU after fine-tuning.

The parameter-efficient variants show comparable performance to full fine-tuning. SegFormer-B5 + LoRA achieves 83.16\% mIoU, slightly improving over the fully fine-tuned version. AdaLoRA variants demonstrate similar trends, with SegFormer-B5 + AdaLoRA reaching 82.81\% mIoU.

\subsubsection{Test Dataset Results}

Table~\ref{tab:test} shows results on the held-out test set. All models exhibit performance degradation compared to the validation results, indicating a distribution shift between the validation and test distributions. SegFormer-B5 maintains the best test mIoU at 61.27\%, followed by SegFormer-B4 at 59.85\%.

Notably, SAM preserves relatively high recall (81.46\%) on the test set despite lower precision, suggesting robustness in detecting landslide regions even under distribution shift. The LoRA and AdaLoRA variants show a consistent behavior, with SegFormer-B5 + AdaLoRA achieving 61.76\% mIoU.

\subsection{Analysis}

\textbf{Precision-Recall Trade-off}: SAM exhibits high recall but lower precision across both validation and test sets, indicating a tendency to over-segment landslide regions. In contrast, SegFormer models achieve more balanced precision-recall trade-offs.

\textbf{Generalization Gap}: The fine-tuned foundation models (e.g., SAM) exhibit noticeable mIoU degradation from validation to test ($\approx$ 16--22 points), indicating potential overfitting or distribution shift in the test split. This gap is observed for both full fine-tuning and parameter-efficient methods.

\textbf{Parameter Efficiency}: LoRA and AdaLoRA reduce trainable parameters by approximately 95\% compared to full fine-tuning while achieving comparable or slightly improved performance. This demonstrates the viability of parameter-efficient adaptation for remote sensing applications.

\textbf{Model Scaling}: Larger model variants (B5 vs B4) consistently outperform smaller ones, suggesting that increased model capacity benefits landslide segmentation, particularly for capturing complex terrain patterns.

\begin{figure}[htbp]
    \centering
        \caption{Qualitative results of landslide segmentation using SegFormer-B4 on the GDCLD test set. Each row shows (from left to right): Input Image, Ground Truth Mask, Predicted Mask, and Overlay with landslide regions highlighted in red.}
    \includegraphics[width=\columnwidth]{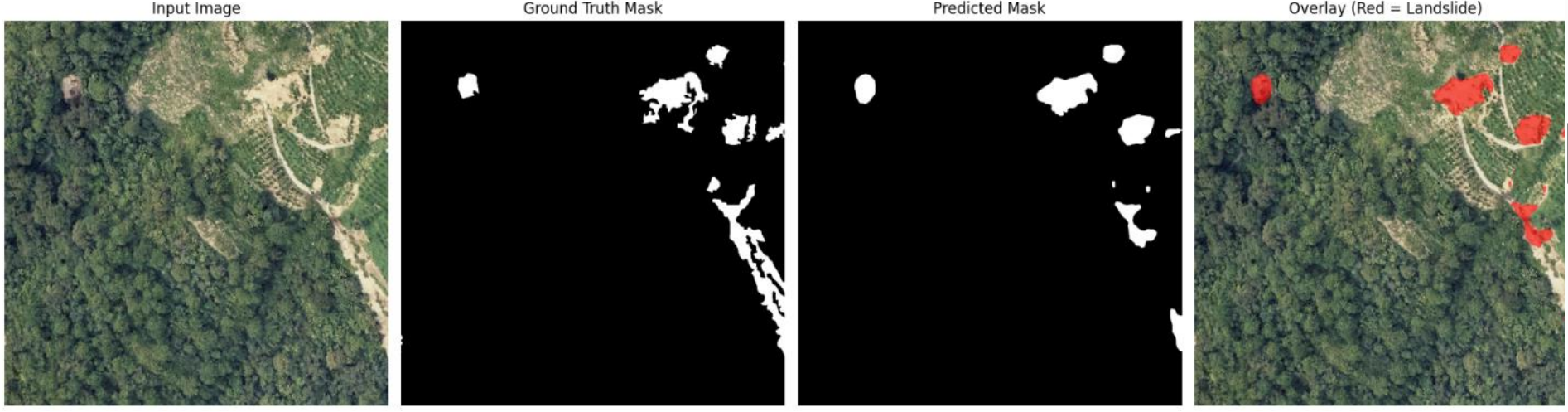}
    
    \vspace{0.05cm}  
    
    \includegraphics[width=\columnwidth]{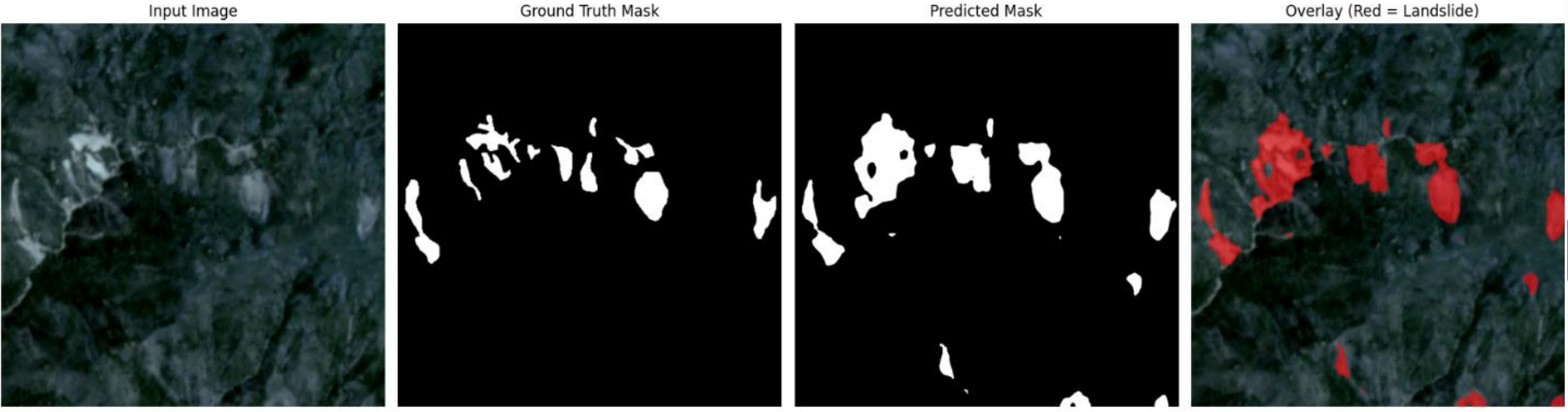}
    
    \vspace{0.05cm}
    
    \includegraphics[width=\columnwidth]{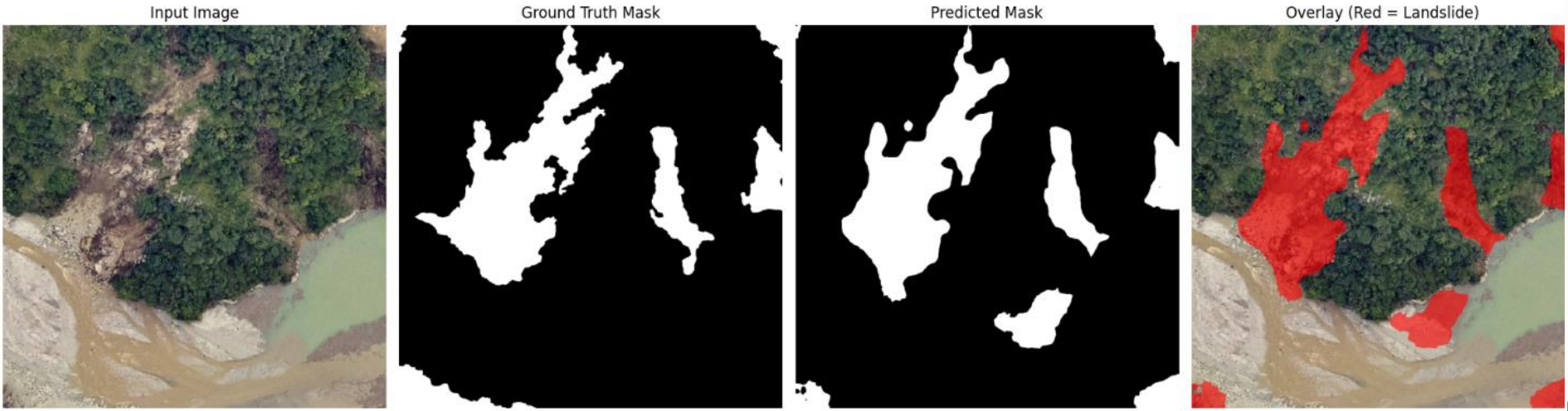}

    \label{fig:landslide_results}
    
\end{figure}

\begin{table}[t]
\centering
\caption{Comparison of results with the GDCLD validation dataset. Models highlighted in green are our fine-tuned pre-trained models and PEFT variants.}
\label{tab:validation}
\resizebox{\columnwidth}{!}{%
\begin{tabular}{@{}llcccc@{}}
\toprule
\textbf{Method} & \textbf{Backbone} & \textbf{Prec.} & \textbf{Rec.} & \textbf{F1} & \textbf{mIoU} \\
\midrule
U-Net & - & 77.05 & 82.01 & 79.54 & 71.07 \\
ResU-Net & ResNet-50 & 78.17 & 86.48 & 82.11 & 71.94 \\
DeepLabV3 & ResNet-50 & 81.27 & 86.96 & 84.02 & 74.61 \\
HRNet & HRNet-48 & 81.88 & 87.21 & 84.46 & 75.19 \\
UPerNet & ViT-B16 & 88.18 & 90.64 & 89.39 & 81.97 \\
SwinU-Net & - & 89.78 & 92.01 & 90.72 & 83.68 \\
SegFormer & MiT-B4 & 91.35 & 91.70 & 91.52 & 85.06 \\
\rowcolor{green!15} SAM & ViT-Base & 79.21 & 89.16 & 83.22 & 73.58 \\
\rowcolor{green!15} SegFormer-B4 & MiT-B4 & 85.27 & 93.76 & 88.91 & 81.17 \\
\rowcolor{green!15} SegFormer-B5 & MiT-B5 & 87.12 & 94.38 & 90.21 & 82.95 \\
\midrule
\rowcolor{green!15} SAM + LoRA & ViT-Base & 79.55 & 88.89 & 83.20 & 73.50 \\
\rowcolor{green!15} SegFormer-B4 + LoRA & MiT-B4 & 85.71 & 94.14 & 89.50 & 80.96 \\
\rowcolor{green!15} SegFormer-B5 + LoRA & MiT-B5 & 87.24 & 94.11 & 90.13 & 83.16 \\
\rowcolor{green!15} SAM + AdaLoRA & ViT-Base & 78.94 & 89.06 & 83.57 & 73.82 \\
\rowcolor{green!15} SegFormer-B4 + AdaLoRA & MiT-B4 & 85.19 & 94.05 & 89.42 & 80.88 \\
\rowcolor{green!15} SegFormer-B5 + AdaLoRA & MiT-B5 & 87.63 & 94.78 & 90.25 & 82.81 \\
\bottomrule
\end{tabular}%
}
\end{table}

\begin{table}[t]
\centering
\caption{Comparison of results with the GDCLD test dataset. Models highlighted in green are our fine-tuned pre-trained models and PEFT variants.}
\label{tab:test}
\resizebox{\columnwidth}{!}{%
\begin{tabular}{@{}llcccc@{}}
\toprule
\textbf{Method} & \textbf{Backbone} & \textbf{Prec.} & \textbf{Rec.} & \textbf{F1} & \textbf{mIoU} \\
\midrule
U-Net & - & 61.69 & 61.22 & 61.45 & 56.09 \\
ResU-Net & ResNet-50 & 66.56 & 64.46 & 65.49 & 57.06 \\
DeepLabV3 & ResNet-50 & 65.26 & 67.75 & 66.48 & 59.73 \\
HRNet & HRNet-48 & 65.52 & 72.03 & 68.62 & 61.79 \\
UPerNet & ViT-B16 & 69.96 & 78.08 & 73.80 & 65.42 \\
SwinU-Net & - & 71.56 & 82.26 & 76.54 & 67.18 \\
SegFormer & MiT-B4 & 77.09 & 87.09 & 81.88 & 72.84 \\
\rowcolor{green!15} SAM & ViT-Base & 59.67 & 81.46 & 64.41 & 57.48 \\
\rowcolor{green!15} SegFormer-B4 & MiT-B4 & 63.95 & 79.18 & 68.47 & 59.85 \\
\rowcolor{green!15} SegFormer-B5 & MiT-B5 & 65.34 & 80.65 & 70.12 & 61.27 \\
\midrule
\rowcolor{green!15} SAM + LoRA & ViT-Base & 59.42 & 81.25 & 64.52 & 57.29 \\
\rowcolor{green!15} SegFormer-B4 + LoRA & MiT-B4 & 64.55 & 78.92 & 68.71 & 59.88 \\
\rowcolor{green!15} SegFormer-B5 + LoRA & MiT-B5 & 65.89 & 80.51 & 70.16 & 61.30 \\
\rowcolor{green!15} SAM + AdaLoRA & ViT-Base & 59.62 & 81.16 & 64.55 & 57.27 \\
\rowcolor{green!15} SegFormer-B4 + AdaLoRA & MiT-B4 & 64.25 & 78.95 & 68.49 & 60.00 \\
\rowcolor{green!15} SegFormer-B5 + AdaLoRA & MiT-B5 & 65.95 & 80.44 & 69.96 & 61.76 \\
\bottomrule
\end{tabular}%
}
\end{table}

\section{Conclusions and Future Work}

This work benchmarked representative CNN-based, transformer-based, and foundation models for coseismic landslide segmentation under consistent training and evaluation settings. The results show that transformer-based architectures achieve strong performance, while parameter-efficient fine-tuning methods substantially reduce training cost with minimal impact on accuracy. Performance degradation on the test set highlights the challenge of distribution shift in real-world scenarios. Future work will focus on improving generalization across datasets, geographic regions, and diverse landslide types.

\small
\bibliographystyle{IEEEtranN}
\bibliography{references}

\end{document}